\documentclass[10pt,twocolumn,letterpaper]{article}

\usepackage{iccv}
\usepackage[utf8]{inputenc}
\usepackage{times}
\usepackage{color}
\usepackage{xcolor}
\usepackage{epsfig}
\usepackage{graphicx}
\usepackage{amsmath}
\usepackage{amssymb}
\newcommand{\mat}[1]{\boldsymbol{#1}}
\renewcommand{\vec}[1]{\boldsymbol{#1}}
\usepackage{gensymb}

\newcommand{\xmark}{x}%

\usepackage{url}

\iccvfinalcopy % *** Uncomment this line for the final submission

\def\iccvPaperID{1743} % *** Enter the ICCV Paper ID here

\ificcvfinal\fi
\begin{document}

%%%%%%%%% TITLE
% Color and depth general visual object tracking benchmark
% Beyond Traditional RGB-D Tracking Benchmarks
% VOT-RGB-D: A Dataset and Benchmark for RGB-D Tracking
\title{\vspace{-0.3cm} CDTB: A Color and Depth Visual Object Tracking Dataset and Benchmark}
%\title{VOT-RGB-D: A Dataset and Benchmark for RGB-D Tracking \cmnt{[MK] how about this: Color and depth general visual object tracking benchmark}}
%\title{VOT-RGB-D: A Dataset and Benchmark for RGB-D Tracking \cmnt{[JK]title proposal 2}}

\author{ Alan Lukežič$^1$, Ugur Kart$^2$, Jani Käpylä$^2$, Ahmed Durmush$^2$, Joni-Kristian Kämäräinen$^2$, \\ Jiří Matas$^3$ and Matej Kristan$^1$ \\
{\small $^1$~Faculty of Computer and Information Science, University of Ljubljana, Slovenia} \\
{\small $^2$~Laboratory of Signal Processing, Tampere University, Finland} \\
{\small $^3$~Faculty of Electrical Engineering, Czech Technical University in Prague, Czech Republic} \\
{\tt\small alan.lukezicfri.uni-lj.si}
}

\maketitle
%\thispagestyle{empty}

%\cmnt{[JK] Do we want to mention something about the results e.g.: "As a striking result, the state-of-the-art RGB trackers are clearly superior to RGB-D trackers indicating that there is yet work to be done in RGB-D tracking to exploit the depth cue more effectively and robustly."}

%%%%%%%%% ABSTRACT
\begin{abstract}
%\cmnt{[MK] IMPORTANT: a crucial downside of our 'Benchmark' is that we evaluate much less RGB-D trackers than PTB. I think we should have s strong argument why this is acceptable. Not a long paragraph, but something we can say concisely.}
We propose a new color-and-depth general visual object tracking benchmark (CDTB). CDTB is recorded by several passive and active RGB-D setups and contains indoor as well as outdoor sequences acquired in direct sunlight. The CDTB dataset is the largest and most diverse dataset for RGB-D tracking, with an order of magnitude larger number of frames than related datasets. The sequences have been carefully recorded to contain significant object pose change, clutter, occlusion, and periods of long-term target absence to enable tracker evaluation under realistic conditions. Sequences are per-frame annotated with 13 visual attributes for detailed analysis. Experiments with RGB and RGB-D trackers show that CDTB is more challenging than previous datasets. State-of-the-art RGB trackers outperform the recent RGB-D trackers, indicating a large gap between the two fields, which has not been detected by the prior benchmarks. Based on the results of the analysis we point out opportunities for future research in RGB-D tracker design. 
\end{abstract}

%%%%%%%%% BODY TEXT
%%%%%%%%%%%%%%%%%%%%%%%%%%%%%%%%%%%%%%%%%%%%%%%%%%%%
\section{Introduction}

\begin{figure}[!th]
\centering
\includegraphics[width=\linewidth]{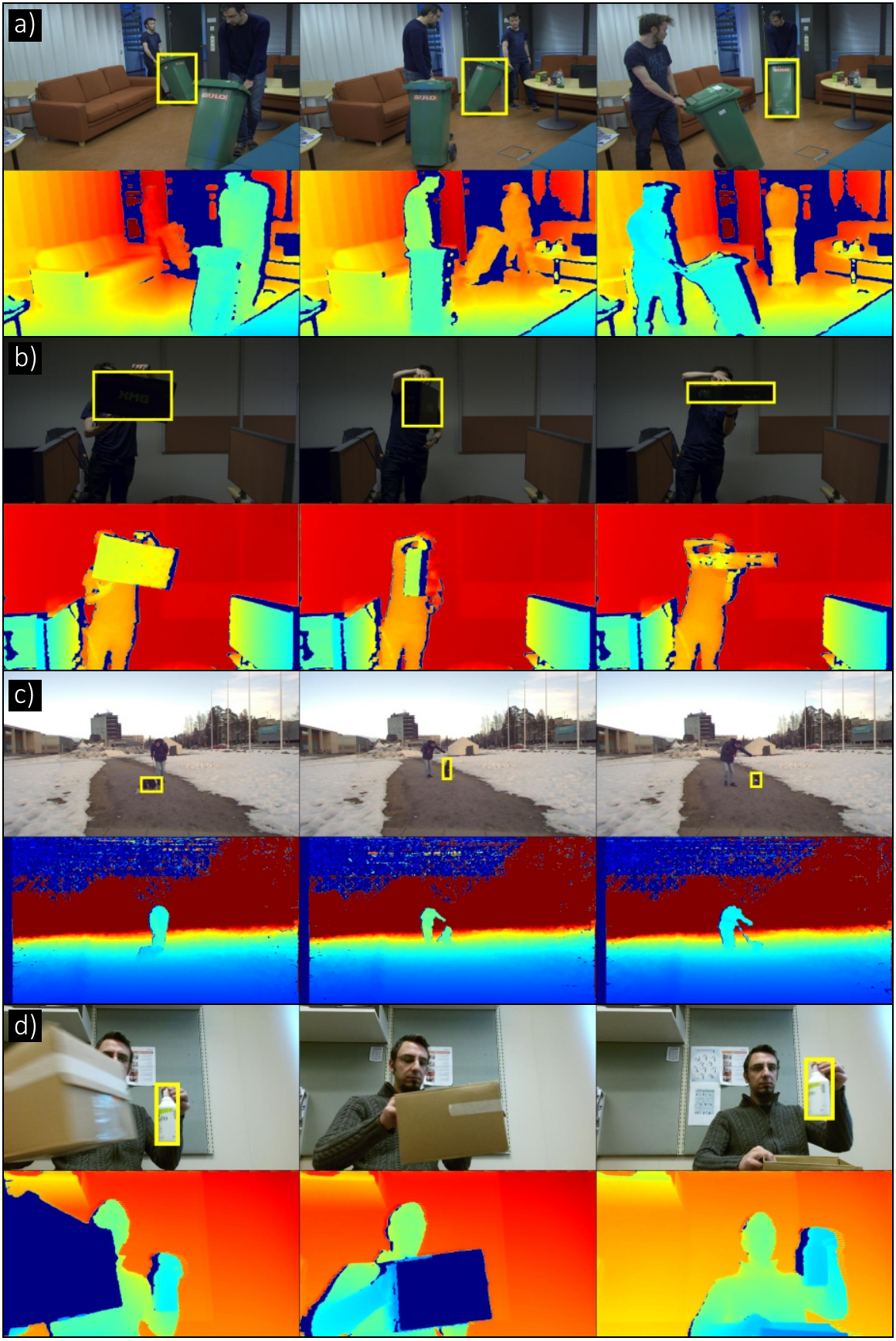} 
\caption{RGB and depth sequences from CDTB. Depth offers a complementary information to color: 
two identical objects are easier to distinguish in depth (a), low illumination scenes (b) are less challenging for trackers if depth information is available, tracking a deformable object in depth simplifies the problem (c) and a sudden significant change in depth is a strong clue for occlusion (d). Sequences (a,b) are captured by a ToF-RGB pair of cameras, (c) by s tereo-camera sensor and (d) by a Kinect sensor.}
\label{fig:first-image}
\end{figure}

Visual object tracking has been enjoying a significant interest of the research community for over several decades due to scientific challenges it presents and its large practical potential. In its most general formulation, it addresses localization of an arbitrary object in all frames of a video, given a single annotation specified in one frame. 
This is a challenging task of self-supervised learning, since a tracker has to localize and carefully adapt to significant target appearance changes, cope with ambient changes, clutter, and detect occlusion and target disappearance. As such, general object trackers cater a range of applications and research challenges like surveillance systems, video editing, sports analytics and autonomous robotics.

%Visual object tracking is the task of localizing an arbitrary object in all frames of a video, given a single annotation specified in the first frame. This is a challenging task of self-supervised learning, since a tracker needs to cope with potentially significant target appearance changes, occlusions, ambient changes, target disappearance, etc. Tracking general objects affords a range of applications from surveillance systems, video editing and sports analytic to robotics.  Because of the scientific challenges and the large practical potential, tracking enjoys a significant interest from the research community. 

Fuelled by emergence of tracking benchmarks~\cite{smeulders_pami_2014,OTB,Kristan2016Pami,Vot2018,Muller_2018_ECCV,uav_benchmark_simulator} that facilitate objective comparison of different approaches, the field has substantially advanced in the last decade. Due to a wide adoption of RGB cameras, the benchmarks have primarily focused on color (RGB) trackers and trackers that combine color and thermal (infrared) modalities~\cite{Vot2014,vot2015,vot2016,vot2017}.

Only recently various depth sensors like RGB-D, time-of-flight (ToF) and LiDAR have become widely accessible. Depth provides an important cue for tracking since it simplifies reasoning about occlusion and offers a better object-to-background separation compared to only color. In addition, depth is a strong cue to acquire object 3D
structure and 3D pose without a prior 3D model, which is crucial in research areas like robotic manipulation~\cite{Buch-2013-icra}. The progress in RGB-D tracking has been boosted by the emergence of RGB-D benchmarks~\cite{princetonrgbd,STC}, but the field significantly lags behind the advancements made in RGB-only tracking. 

One reason for the RGB -- RGB-D general object tracking performance gap is that existing RGB-D benchmarks~\cite{princetonrgbd,STC} are less challenging than their RGB counterparts. The sequences are relatively short from the perspective of practical applications, the objects never leave and re-enter the field of view, they undergo only short-term occlusions and rarely significantly rotate away from the camera. The datasets are recorded indoor only with Kinect-like sensors which prohibits generalization of the results to general outdoor setups. These constraints were crucial for early development of the field, but further boosts require a more challenging benchmark, which is the topic of this paper.   
 
In this work we propose a new color-and-depth tracking benchmark (CDTB) that makes several contributions to the field of general object RGB-D tracking. (i) The CDBT dataset is recorded by several color-and-depth sensors to capture a wide range of realistic depth signals. (ii) The sequences are recorded indoor as well as outdoor to extend the domain of tracking setups. (iii) The dataset contains significant object pose changes to encompass realistic depth appearance variability. (iv) The objects are occluded or leave the field of view for longer duration to emphasize the importance of trackers being able to report target loss and perform re-detection. (v) We compare several state-of-the-art RGB-D trackers as well as state-of-the-art RGB trackers and their RGB-D extensions.
Examples of CDTB dataset are shown in Figure~\ref{fig:first-image}.
%\cmnt{[JK] Do we want to mention something about the results e.g.: "As a striking result, the state-of-the-art RGB trackers are clearly superior to RGB-D trackers indicating that there is yet work to be done in RGB-D tracking to exploit the depth cue more effectively and robustly."}

The reminder of the paper is structured as follows. Section~\ref{sec:trackers} summarizes the related work, Section~\ref{sec:dataset} details the acquisition and properties of the dataset, Section~\ref{sec:performance-measures} summarizes the performance measures, Section~\ref{sec:experiments} reports experimental results and Section~\ref{sec:conclusion} concludes the paper.

%The benchmarks have primarily focused on RGB and thermal sequences due to their 

%The benchmarks have primarily focused on short-term RGB tracking~\cite{?,?,?}, i.e., on trackers that assume target is always present in the frame and are not required to perform re-detection, and few benchmarks have considered tracking over long sequences with targets often leaving and re-entering the field of view~\cite{valmadre_lt_benchmark,Lukezic_arxiv_lt_bench,Vot2018}.
 
%\paragraph{Contributions} of our work are

%%%%%%%%%%%%%%%%%%%%%%%%%%%%%%%%%%%%%%%%%%%%%%%%%%%%
\section{Related work}\label{sec:trackers}
%In this section we briefly discuss the available RGB-D tracking benchmark datasets and trackers.

%
%
%\subsection{RGB-D Benchmarks}

%There are two publicly available benchmarks for RGB-D tracking: {\em Princeton Tracking Benchmark} (PTB)~\cite{princetonRGB-D} and {\em Spatio-Temporal Consistency} (STC) dataset~\cite{STC}. ...and evaluation is made by uploading the tracking results to the online system provided by the authors of~\cite{princetonRGB-D}. 
%The PTB sequences have several shortcomings.
%At first, many sequences share the same objects and background. For example, more than half of the sequences are pedestrian tracking.
%Secondly, the sequences have synchronization errors between the RGB and D frames (14\% in total) and
%their alignment (8\%). Bibi et al~\cite{Bibi3D} provide corrected
%sequences at their project Web page and report results for the both
%original and corrected data. %The average length of the PTB
%videos is \cmnt{[JK] XX.X seconds (YYYY frames)}.
%[MK] This should go into the benchmark comparison section

\paragraph{RGB-D Benchmarks.}
% I AM STILL EDITING, DO NOT TOUCH //JONI
The diversity of the RGB-D datasets is limited compared to those in RGB tracking. Many
of the datasets are application specific, e.g.,
{\em pedestrian tracking} or {\em hand tracking}.
For example,
Ess~\etal~\cite{Ess-2008-cvpr} provide five
3D bounding box annotated sequences captured by a
calibrated stereo-pair,
the RGB-D People Dataset~\cite{Spinello-2011-iros} contains
a single sequence of pedestrians in a hallway captured by
a static RGB-D camera and Stanford Office~\cite{choi_pami13} contains 17 sequences with a static and one with a moving
Kinect. Garcia-Hernando~\etal~\cite{Garcia-Hernando-2018-cvpr}
introduce an RGB-D dataset for hand tracking and action recognition.
Another important application
field for RGB-D cameras is robotics, but
here datasets are often small and the main objective is
real-time model-based 3D pose estimation.
For example, the
RGB-D Object Pose Tracking Dataset~\cite{choi13iros_rgbdtracking}
contains 4 synthetic and 2 real RGB-D image sequences to benchmark
visual tracking and 6-DoF pose estimation. Generating synthetic data
has become popular due to requirements of large training sets
for deep methods~\cite{Richter-2016-eccv}, but
it is unclear how well these predict 
real world performance.

Only two datasets are dedicated to general object tracking. The most popular is Princeton Tracking Benchmark (PTB)~\cite{princetonrgbd}, which contains 100 RGB-D video sequences of rigid and nonrigid objects recorded with Kinect. The choice of sensor constrains the dataset to only indoor scenarios. The dataset diversity is further reduced since many sequences share the same tracked objects and the background. More than half of the sequences are people tracking. The sequences are annotated by five global attributes. The RGB and depth channels are poorly calibrated. In approximately 14\% of sequences the RGB and D channels are not synchronized and approximately 8\% are miss-aligned.
The calibration issues were addressed by Bibi et al~\cite{Bibi3D} who published a corrected dataset. PTB addresses long-term tracking, in which the tracker has to detect target loss and perform re-detection. The dataset thus contains several full occlusions, but the target never leaves and re-enters the field of view, thus limiting the evaluation capabilities of re-detecting trackers. Performance is evaluated as the percentage of  frames in which the bounding box predicted by tracker exceeds a $0.5$ overlap with the ground truth. The overlap is artificially set to 1 when the tracker accurately predicts target absence. Recent work in long-term tracker performance evaluation~\cite{valmadre_lt_benchmark,Lukezic_arxiv_lt_bench} argue against using a single threshold and~\cite{Lukezic_arxiv_lt_bench} further show reduced interpretation strength of the measure used in PTB.
 
The Spatio-Temporal Consistency dataset (STC)~\cite{STC} was recently proposed to address the drawbacks of PTB. The dataset is recorded by Asus Xtion RGB-D sensor, which also constrains the dataset to only indoor scenarios and a few low-light outside scenarios, but care has been taken to increase the sequence diversity. The dataset is smaller than PTB, containing only 36 sequences, but annotated by thirteen global attributes. STC addresses short-term tracking scenario, i.e., trackers are not required to perform re-detection. 
Thus the sequences are relatively short and the short-term performance evaluation methodology is used. This makes the dataset inappropriate for evaluating trackers useful in many practical setups, in which target loss detection and redetection are crucial capabilities.  

%STC is proposed to complement the PTB with more diverse and error-free sequences. The sequences are recorded using the Asus Xtion RGB-D
%sensor. The dataset contains 36 sequences of various objects and
%settings and for detailed evaluation the sequences are annotated
%with $10$ attributes; based on computational statistics (CS):
%Illumination variation, Depth variation, Scale variation,
%Color distribution variation, Depth distribution variation, Surrounding depth clutter and Surrounding color clutter; and based on
%human annotations (HA): Background color camouflages,
%Background shape camouflages and Partial occlusion. 

%The average length
%of the STC sequences is \cmnt{[JK] XX.X seconds (YYYY frames)}. -- [MK] This should go into the benchmark comparison section.

\paragraph{RGB Trackers.}
Recent years have seen a surge in Short-term Trackers (ST) and especially Discriminative Correlation Filter (DCF) based approaches have been popular due to their mathematical simplicity and elegance. In their seminal paper, Bolme~\etal~\cite{Bolme-2010-cvpr} proposed using DCF for visual object tracking. 
Henriques~\etal~\cite{Henriques_KCF} proposed an efficient training method by exploiting the properties of circular convolution. 
Lukezic~\etal~\cite{csr} and Galoogahi~\etal~\cite{Galoogahi-2015-cvpr}  proposed a mechanism to handle boundary problems and segmentation-based DCF constraints have been introduced in~\cite{csr}. 
Danelljan~\etal~\cite{ECO} used a factorized convolution operator and achieved excellent scores on well-known benchmarks. 

As a natural extension of the ST, Long-term Trackers (LT) have been proposed~\cite{TLD} where the tracking is decomposed into short-term tracking and long-term detection. 
Lukezic~\etal proposed a fully-correlational LT~\cite{Fucolot} by storing multiple correlation filters that are trained at different time scales. 
Zhang~\etal~\cite{MBMD} used deep regression and verification networks and they achieved the top rank in VOT-LT 2018~\cite{Vot2018}.
Despite being published as an ST, MDNet~\cite{MDNet} has proven itself as an efficient LT. MDNet uses discriminatively trained Convolutional Neural Networks(CNN) and won the VOT 2015 challenge~\cite{vot2015}.

\paragraph{RGB-D Trackers.}  
Compared to RGB trackers, the body of literature on RGB-D trackers is rather limited which can be attributed to the lack of available datasets until recently. In 2013, the publication of PTB~\cite{princetonrgbd} ignited the interest in the field and there have been numerous attempts by adopting different approaches. The authors of PTB have proposed multiple baseline trackers which use different combinations of HOG~\cite{HOG}, optical flow and point clouds. As a part of particle filter tracker family, Meshgi~\etal~\cite{MESHGI_OAPF} proposed a particle filter framework with occlusion awareness using a latent occlusion flag. They pre-emptively predict the occlusions, expand the search area in case of occlusions. Bibi~\etal~\cite{Bibi3D} represented the target by sparse, part-based 3-D cuboids while adopting particle filter as their motion model. Hannuna~\etal~\cite{Hannuna2016}, An~\etal~\cite{DLST} and Camplani~\etal~\cite{dskcf_bmvc} extended the Kernelized Correlation Filter (KCF) RGB tracker~\cite{Henriques_KCF} by adding the depth channel. Hannuna~\etal and Camplani~\etal proposed a fast depth image segmentation which is later used for scale, shape analysis and occlusion handling. An~\etal proposed a framework where the tracking problem is divided into detection, learning and segmentation. To use depth inherently in DCF formulation, Kart~\etal~\cite{DMDCF} adopted Gaussian foreground masks on depth images in CSRDCF~\cite{csr} training. They later extended their work by using a graph cut method with color and depth priors for the foreground mask segmentation~\cite{Kart_ECCVW} and more recently proposed a view-specific DCF using object's 3D structure based masks~\cite{OTR}. Liu~\etal~\cite{ca3dms} proposed a 3D mean-shift tracker with occlusion handling. Xiao~\etal~\cite{STC} introduced a two-layered representation of the target by adopting a spatio-temporal consistency constraints.

\section{Color and depth tracking dataset}\label{sec:dataset}
 
% 58 RGB+ToF, 12 Kinect, 10 Stereo RGB 
 
%\subsection{Acquisition hardware}

\begin{figure}[t]
    \includegraphics[width=0.49\linewidth]{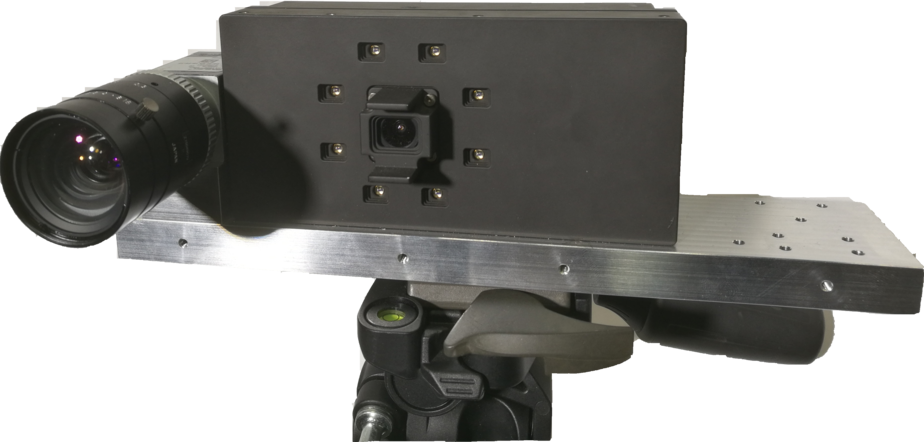} \hfill
    \includegraphics[width=0.45\linewidth]{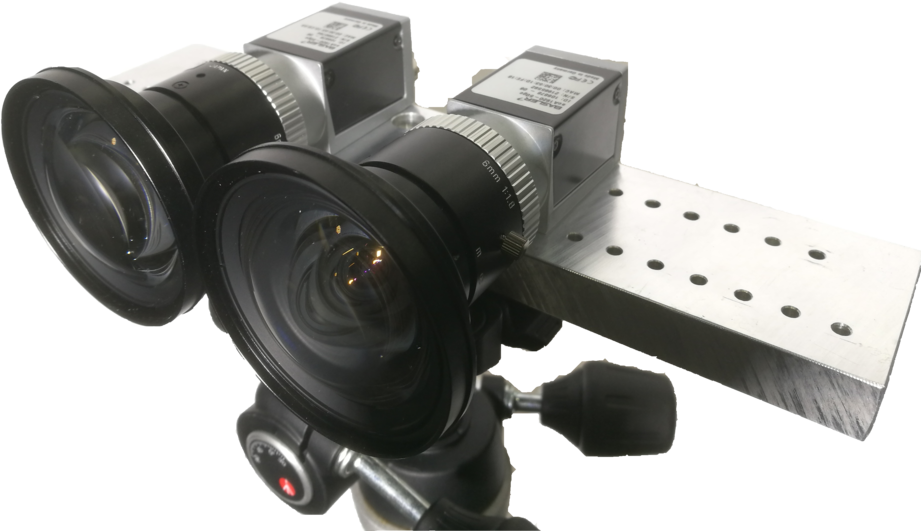}
    \caption{Two of the three sensors used in dataset acquisition:
    ToF-RGB-pair (left) and a stereo-cameras pair (right).
    The third sensor, Kinect v2, is standard.
    \label{fig:setup}} \end{figure}

We used several RGB-D acquisition setups to increase the dataset diversity in terms of acquisition hardware. This allowed unconstrained indoor as well as outdoor sequence acquisition, thus diversifying the dataset and broaden the scope of realistic scenarios. The following three acquisition setups were used: (i) RGB-D sensor (Kinect), (ii) time-of-flight (ToF)-RGB pair and (iii) stereo cameras pair (Figure~\ref{fig:setup}). The setups are described in the following.
%\cmnt{[AL] Do we want to say how many sequences were taken using which sensor?} 

\paragraph{RGB-D Sensor} sequences were captured with
a Kinect v2 that outputs 24-bit $1920\times 1080$ RGB images (8-bit per color channel) and $512\times 424$ 32-bit floating point depth images with an average frame rate of $30$~fps. JPEG compression is applied to RGB frames while depth data is converted into 16-bit unsigned integer and saved in PNG format. The RGB and depth images are synchronized internally and no further synchronization was required.

\paragraph{ToF-RGB pair} consists of Basler tof640-20gm time-of-flight and Basler acA1920-50gc color cameras. The ToF camera has 640x480pix resolution and maximum 20 fps frame rate whereas color camera has 1920x1200pix resolution and 50 fps maximum frame rate at full resolution. Both cameras can be triggered externally using the I/O's of the cameras for external synchronisation. The cameras were mounted on a high precision CNC-machined aluminium base in a way that the baseline of the cameras are 75.2mm and camera sensor center points are on the same level. The TOF camera has built in optics with 57$^\circ\times$43$^\circ$ (HxV) field-of-view. The color camera was equipped with a 12mm focal length lens (VS-1214H1), which has 56.9$^\circ\times$44$^\circ$ (HxV) field-of-view for 1" sensors, to match the field-of-view of the ToF camera. The cameras were synchronised by an external triggering device at the rate of 20~fps. The color camera output was 8-bit raw Bayer images whereas ToF camera output was 16-bit depth images. The raw Bayer images were later debayered to 24-bit RGB images (8-bit per color channel).

%was set up on a custom-made multipurpose camera rig with multiple camera mounts and can be itself attached to a standard camera tripod (Figure~\ref{fig:setup}). A commercial-grade Basler tof640-20gm time-of-flight camera that captures $640\times 480$ 16-bit depth images at rate of 20~fps was used for depth acquisition. The RGB images were captured by Basler Aca 1920-50gc that outputs $1920\times 1200$  24-bit color images (8-bit per color channel) at 50fps. The camera was equipped with a VS Technology lens vs-1214H1 of 12~mm focal length. The ToF-RGB baseline was set to $75.2$~mm. The setup was equipped by an external trigger and a vendor provided library was used to capture RGB and depth images simultaneously. The trigger was set to capture frames at 20 fps.

\paragraph{Stereo-cameras pair} is composed of two Basler acA1920-50gc color cameras which are mounted on a high precision machined aluminium base with 70mm baseline. The cameras were equipped with 6mm focal length lenses (VS-0618H1) with 98.5$^\circ\times$77.9$^\circ$ (HxV) field-of-view for 1" sensors. The cameras were synchronised by an external triggering device at the rate of 40~fps at full resolution. The camera outputs were 8-bit raw Bayer images which were later Bayer demosaiced to 24-bit RGB images (8-bit per color channel). A semi-global block matching algorithm~\cite{Hirschmuller2005} was applied to the rectified stereo images and converted to metric depth values using the camera calibration parameters.

%was composed of two Basler Aca 1920-50gc cameras, each outputting $1920\times 1200$ 24-bit RGB images (8-bit per color channel) at 40fps. The stereo rig baseline was 70.0~mm. To cope with parallax, VS Technology wide angle lenses vs-0618H1 of 6~mm focal length were used. A custom-made sync external trigger using a CC320 Trigger Timing Controller was used to capture both images in sync. A semi-global block matching algorithm~\cite{pami_semi_global_block_matching} was applied to the rectified stereo images and converted to metric depth values using the camera calibration parameters.

\subsection{RGB and Depth Image Alignment}%{Calibration}

All three acquisition setups were calibrated using the Caltech Camera Calibration Toolbox\footnote{{\scriptsize \url{http://www.vision.caltech.edu/bouguetj/calib_doc}}}
with standard modifications to cope with image pairs of different resolution
for the RGB-D sensor and ToF-RGB-pair setups.
% NOTE: code we used: https://github.com/balcilar/Calibration-Under_Different-Resolution
%The toolbox checkerboard pattern was printed on a large ($1.0~m\times 1.0~m$) plastic plate and tens of images were captured for each setup. 
The calibration provides the external camera parameters, {\em rotation matrix} $\mat{R}_{3\times 3}$ and {\em translation vector} $\vec{t}_{3\times 1}$, and the intrinsic camera parameters, {\em focal length} $\vec{f}_{2\times 1}$, {\em principal point} $\vec{c}_{2\times 1}$, {\em skew $\alpha$} and  lens distortion coefficients $\vec{k}_{5\times 1}$. The forward projection is defined by~\cite{HZ}
\begin{equation}
\vec{m} = \mathcal{P}(\vec{x})=(\mathcal{P}_c \circ \mathcal{R})(d),
%\mathcal{P}_d (\vec{x}_c)=\mathcal{K} \circ \mathcal{D} \circ %\hat{\mathcal{\nu}} (\vec{x}_c)
\label{eq:projection}
\end{equation}
where $\vec{x} = (x,y,z)^T$ is the scene point in world coordinates, $\vec{m}$ is the projected point in image coordinates and $d=I_{depth}(\vec{m})$ is the depth. 
$\mathcal{R}$ is a rigid Euclidean transformation, $\vec{x}_c = \mathcal{R}(\vec{x})$,
defined by $\mat{R}$ and $\vec{t}$, and $\mathcal{P}_c$ is the
intrinsic operation
$\mathcal{P}_c(\vec{x}_c)=(\mathcal{K} \circ \mathcal{D} \circ \hat{\mathcal{\nu}})(\vec{x}_c)$ of the perspective division
operation $\hat{\mathcal{\nu}}$, distortion operation
$\mathcal{D}$ using $\vec{k}$ and the affine mapping
$\mathcal{K}$ of $\vec{f}$ and $\alpha$.

%
%
%\subsection{Video Synchronization}
%RGB-D Sensor synchronization was not needed as the Kinect
%SDK provides simultaneous capture of RGB and depth images.\cmnt{[UK] We used libfreenect2}
%
%ToF-RGB-pair synchronization was done by using the vendor
%provided library to capture the depth and RGB images simultaneously
%
%For the stereo-pair setup we constructed a  custom-made sync signal
%trigger using a CC320 Trigger Timing Controller. The Basler
%ToF sensor and RGB cameras provide Ethernet interface for
%data transfer and Kinect uses USB.

%
%
%\subsection{RGB and Depth Image Alignment}

The depth images of RGB-D Sensor and ToF-RGB pair were per-pixel aligned to the RGB images as follows. 
A 3D point corresponding to each pixel in the calibrated depth image was computed using the inverse of the equation (\ref{eq:projection}) as $\vec{x} = \mathcal{P}^{-1}(\vec{m},d)$. 
These points were projected to the RGB image by a linear interpolation model. For further studies we provide the original data and calibration parameters upon request.

%Each pixel $d=I_{depth}(x,y)$ of a calibrated depth image maps to a unique 3D point by the inverse of the equation (\ref{eq:projection}) as $\vec{x} = \mathcal{P}^{-1}(\vec{m},d)$.
%The points can be forward mapped to the RGB plane by the forward projection (\ref{eq:projection}) and each RGB pixel
% assigned a depth value by nearest-neighbor interpolation. For optimal results occlusions should be detected and removed, for example,
% using a {\em soft Z-buffer}~\cite{SoftZBuffer}, but since occlusions were minor in our case we did simple interpolation. For further studies
% we also provide the original data and calibration parameters on request.

%
%
%\subsection{Stereo Reconstruction}
%For stereo-pair depth computation we used the Matlab implementation
%of the popular semi-global block matching algorithm for
%calibrated and rectified stereo-pair images~\cite{MatlabStereoAlg2}. 
%The disparities were converted to metric depth values.

%
%
\subsection{Sequence Annotation}  \label{sec:sequence-annotation}

The VOT Aibu image sequence annotator\footnote{{\scriptsize \url{https://github.com/votchallenge/aibu}}} was used to manually annotate the targets by axis-aligned bounding boxes. The bounding boxes were placed following the VOT~\cite{Vot2014} definition by maximizing the number of target pixels within the bounding box and minimizing their number outside the bounding box. All bounding boxes were checked by several annotators for quality control.

%Track bounding boxes were annotated using {\em Aibo image sequence annotator} provided by the VOT Challenge organizers. The tool provides a user interface where bounding boxes are annotated sparsely and the system interpolates intermediate bounding boxes. If interpolation is inaccurate the user can adjust the interpolated boxes by annotating more boxes.

%All sequences were annotated per-frame with thirteen attributes: (i) target out-of-view, (ii) full occlusion, (iii) partial occlusion, (iv) out-of-plane rotation, (v) fast motion, (vi) similar objects, (vii) target size change, (viii) aspect ratio change, (ix) deformable target, (x) reflective target, (xi) depth change and (xii) dark scene. Frames not annotated with any of the first twelve attributes are annotated by (xiii) unassigned.

All sequences were annotated per-frame with thirteen attributes. The following attributes were manually annotated: 
(i)~target out-of-view, 
(ii)~full occlusion, 
(iii)~partial occlusion, 
(iv)~out-of-plane rotation, 
(v)~similar objects, 
(vi)~deformable target,
(vii)~reflective target and
(viii)~dark scene.
The attribute (ix)~fast motion was assigned to a frame in which the target center moves by at least 30\% of its size in consecutive frames, 
(x)~target size change was assigned when the ratio between maximum and minimum target size in 21 consecutive frames was larger than 1.5 and 
(xi)~aspect ratio change was assigned when the ratio between the maximum and minimum aspect (i.e., width / height) within 21 consecutive frames was larger than 1.5.
The attribute (xii)~depth change was assigned when the ratio between maximum and minimum of median of depth within target region in 21 consecutive frames was larger than 1.5.
Frames not annotated with any of the first twelve attributes were annotated as (xiii)~unassigned.

%%%%%%%%%%%%%%%%%%%%%%%%%%%%%%%%%%%%%%%%%%%%%%%%%%%%%
\section{Performance Evaluation Measures}\label{sec:performance-measures}

Tracker evaluation in a long-term tracking scenario in which targets may disappear/re-appear, requires measuring the localization accuracy, as well as re-detection capability and ability to report that target is not visible. 
To this end we adopt the recently proposed long-term tracking evaluation protocol from~\cite{Lukezic_arxiv_lt_bench}, which is used in the VOT2018 long-term challenge~\cite{Vot2018}. The tracker is initialized in the first frame and left to run until the end of the sequence without intervention. 

The implemented performance measures are tracking precision ($Pr$) and tracking recall ($Re$) from~\cite{Lukezic_arxiv_lt_bench}.
Tracking precision measures the accuracy of target localization when deemed visible, while tracking recall measures the accuracy of classifying frames with target visible. The two measures are combined into F-measure, which is the primary measure. In the following we briefly present how the measures are calculated. For details and derivation we refer the reader to~\cite{Lukezic_arxiv_lt_bench}.

We denote $G_t$ as a ground-truth target pose and $A_{t}(\tau_{\theta})$ as a pose prediction given by a tracker at frame $t$. The evaluation protocol requires that the tracker reports a confidence value besides the pose prediction.
The confidence of the tracker in frame $t$ is denoted as $\theta_t$ while confidence threshold is denoted as $\tau_{\theta}$.
If the target is not visible in frame $t$, then ground-truth is an empty set i.e., $G_t = \emptyset$.
Similarly, if tracker does not report the prediction or if confidence score is below the confidence threshold, i.e., $\theta_t < \tau_{\theta}$, then the output is an empty set $A_{t}(\tau_{\theta}) = \emptyset$.

From the object detection literature, when intersection-over-union between the tracker prediction and ground-truth $\Omega (A_t(\tau_{\theta}), G_t)$, exceeds overlap threshold $\tau_{\Omega}$, the prediction is considered as correct.
This definition of correct prediction highly depends on the minimal overlap threshold $\tau_{\Omega}$. 
The problem is in~\cite{Lukezic_arxiv_lt_bench} addressed by integrating tracking precision and recall over all possible overlap thresholds which results in the following measures
\begin{equation}  \label{eq:precision}
    Pr(\tau_{\theta}) = \frac{1}{N_p} \sum_{t\in \{ t: A_t(\tau_{\theta})\neq \emptyset \}} \Omega (A_t(\tau_{\theta}), G_t),
\end{equation}
\begin{equation}  \label{eq:recall}
    Re(\tau_{\theta}) = \frac{1}{N_g} \sum_{t\in \{ t: G_t\neq \emptyset \}} \Omega (A_t(\tau_{\theta}), G_t),
\end{equation}
where $N_g$ represents number of frames where target is visible, i.e., $G_t \neq 0$ and $N_p$ is number of frames where tracker made a prediction, i.e., $A_t(\tau_{\theta}) \neq \emptyset$.
Tracking precision and recall are combined into a single score by computing tracking F-measure
\begin{equation}  \label{eq:f-measure}
    F(\tau_{\theta}) = \frac{2 Re(\tau_{\theta}) Pr(\tau_{\theta})}{Re(\tau_{\theta}) + Pr(\tau_{\theta})} \enspace .
\end{equation}
Tracking performance is visualized on precision-recall and F-measure plots by computing scores for all confidence thresholds $\tau_{\theta}$.
The highest F-measure on the F-measure plot represents the optimal confidence threshold and it is used for ranking trackers. 
This process also does not require manual threshold setting for each tracker separately. 

The performance measures are directly extended to per-attribute analysis. In particular, the tracking Precision, Recall and F-measure are computed from predictions on the frames corresponding to a particular attribute.

%%%%%%%%%%%%%%%%%%%%%%%%%%%%%%%%%%%%%%%%%%%%%%%%%%%%
\section{Experiments}\label{sec:experiments}

This section presents experimental results on the CDTB dataset. Section~\ref{sec:tested-trackers} summarizes the list of tested trackers, Section~\ref{sec:benchmarks-comparison} compares the CDTB dataset with most related datasets, Section~\ref{sec:baseline-experiment} reports overall tracking performance and Section~\ref{sec:per-attribute-experiment} reports per-attribute performance.
 
\begin{table*}[t]
\begin{center}
\caption{
Comparison of CDTB with related benchmarks in the number of RGB-D devices used for acquisition (N$_\mathrm{HW}$), presence of indoor and outdoor sequences (In/Out), per-frame attribute annotation (Per-frame), number of attributes (N$_\mathrm{atr}$), number of sequences (N$_\mathrm{seq}$), total number of frames (N$_\mathrm{frm}$) average sequence length (N$_\mathrm{avg}$), number of frames with target not visible (N$_\mathrm{out}$), number of target disappearances (N$_\mathrm{dis}$), average length of target absence period (N$_\mathrm{avgout}$), number of times a target rotates away from the camera by at least 180\degree (N$_\mathrm{rot}$), average number of target rotations per sequence (N$_\mathrm{seqrot}$) and tracking performance under the PTB protocol ($\Omega_{0.5}$).  
%Properties of the proposed CDTB and the two currently available RGB-D datasets STC and PTB.
%The column N$_\mathrm{HW}$ corresponds to the number of different sensors used to capture the dataset.
%In and Out denote indoor and outdoor sequences, respectively, Per-frame represents per-frame attribute annotation and N$_\mathrm{atr}$ denotes number of attributes. 
%General properties of the dataset size are number of sequences (N$_\mathrm{seq}$) and total and average number of frames (N$_\mathrm{frm}$ and N$_\mathrm{avg}$).
%Total number of frames where target is not visible is denoted as N$_\mathrm{out}$, while total number of target disappearances is denoted as N$_\mathrm{dis}$. 
%The column N$_\mathrm{avgout}$ represents average length of the period when target is not visible.
%Number of target rotations for at least 180\degree is denoted as N$_\mathrm{rot}$, while N$_\mathrm{seqrot}$ denotes average number of target rotations per-sequence. 
%Average performance of the selected trackers, calculated as percentage of successfully tracked frames at overlap threshold 0.5, is denoted as $\Omega_{0.5}$.
}
\label{tab:datasets-comparison}
\scalebox{.83}{
\begin{tabular}{lcccccrrrrrrrrr}
%\toprule
\hline
Dataset & N$_\mathrm{HW}$ & In & Out & Per-frame & N$_\mathrm{atr}$ & N$_\mathrm{seq}$ & N$_\mathrm{frm}$ & N$_\mathrm{avg}$ &   N$_\mathrm{out}$ & N$_\mathrm{avgout}$ & N$_\mathrm{dis}$ & N$_\mathrm{rot}$ &  N$_\mathrm{seqrot}$ & $\Omega_{0.5}$ \\
%\midrule
\hline
CDTB & 3 & \checkmark & \checkmark &  \checkmark & 13 & 80 & 101,956 & 1,274 & 10,656 & 56.4 & 189 & 358 & 4.5 & 0.316 \\
STC~\cite{STC} & 1 & \checkmark & \checkmark & \checkmark & 12 &  36 & 9,195 & 255 & 0 & 0 & 0 & 30 & 0.8 & 0.530 \\
PTB~\cite{princetonrgbd} & 1 & \checkmark & \xmark & \xmark & 5 & 95 & 20,332 & 214 & 846 & 6.3 & 134 & 83 & 0.9 & 0.749 \\
%\bottomrule
\hline
\end{tabular}
}
\end{center}
\end{table*}

\subsection{Tested Trackers}  \label{sec:tested-trackers}

The following 16 trackers were chosen for evaluation. We tested (i) RGB baseline and state-of-the-art short-term correlation and deep trackers (KCF~\cite{Henriques_KCF}, NCC~\cite{Vot2013}, BACF~\cite{Galoogahi_2017_ICCV}, CSRDCF~\cite{csr}, SiamFC~\cite{bertinetto2016fully}, ECOhc~\cite{ECO}, ECO~\cite{ECO} and MDNet~\cite{MDNet}), (ii) RGB state-of-the-art long-term trackers (TLD~\cite{TLD}, FuCoLoT~\cite{Fucolot} and MBMD~\cite{MBMD}) and (iii) RGB-D state-of-the-art trackers (OTR~\cite{OTR} and Ca3dMS~\cite{ca3dms}). 
Additionally, the following RGB trackers have been modified to use depth information: ECOhc-D~\cite{Kart_ECCVW}, CSRDCF-D~\cite{Kart_ECCVW} and KCF-D\footnote{KCF-D is modified by using depth as a feature channel in a correlation filter. %Target is localized by finding position of the maximum on the averaged correlation response
}.

\subsection{Comparison with Existing Benchmarks}  \label{sec:benchmarks-comparison}

Table~\ref{tab:datasets-comparison} compares the properties of CDTB with the two currently available datasets,  PTB~\cite{princetonrgbd} and STC~\cite{STC}. CDTB is the only dataset that contains sequences
captured with several devices in indoor and outdoor tracking scenes. STC~\cite{STC} does in fact contain a few outdoor sequences, but these are confined to scenes without direct sunlight due to infra-red-based depth acquisition. The number of attributes is comparable to STC and much higher than PTB.
%, but CDTB is the only dataset with per-frame attribute annotation, which affords a detailed analysis at reduced attribute crosstalk~\cite{cehovin_beyondstandadrdbecnharmsk_iccv}. 
The number of sequences (N$_\mathrm{seq}$) is comparable to the currently largest dataset PTB, but CDTB exceeds the related datasets by an order of magnitude in the number of frames (N$_\mathrm{frm}$). 
In fact, the average sequence of CDTB is approximately six times longer than in related datasets (N$_\mathrm{avg}$), which affords a more accurate evaluation of long-term tracking properties. 

A crucial tracker property required in many practical applications is target absence detection and target re-detection. STC lacks these events. The number of target disappearances followed by re-appearance in CDTB is comparable to PTB, but the disappearance periods (N$_\mathrm{out}$) are much longer in CDTB. The average period of target absent (N$_\mathrm{avgout}$) in PTB is approximately 6 frames, which means that only short-term occlusions are present. The average period of target absent in CDTB is nearly ten times larger, which allows tracker evaluation under much more challenging and realistic conditions.

Pose changes are much more frequent in CDTB than in the other two datasets. 
For example, the target undergoes a 180 degree out-of-plane rotation less than once per sequence in PTB and STC (N$_\mathrm{seqrot}$). Since CDTB captures more dynamic and realistic scenarios, the target undergoes such pose change nearly 5 times per sequence.
  
The level of appearance change, realism, disappearances and sequence lengths result in a much more challenging dataset that allows performance evaluation under more realistic conditions than STC and PTB. 
%To quantify this, we have identified trackers evaluated in PTB and STC benchmarks and evaluated these on CDTB. 
To quantify this, we evaluated trackers Ca3dMS, CSR-D and OTR on the three datasets and averaged their results. 
The trackers were evaluated on STC and CDTB using the PTB performance measure, since PTB does not provide ground truth bounding boxes for public evaluation.

Table~\ref{tab:datasets-comparison} shows that the trackers achieve the highest performance on PTB, making it least challenging. 
The performance drops on STC, which supports the challenging small dataset diversity paradigm promoted in~\cite{STC}. 
The performance further significantly drops on CDTB, which confirms that this dataset is the most challenging among the three.

\subsection{Overall Tracking Performance}  \label{sec:baseline-experiment}

%\cmnt{[MK] We would like to argue that: (1) RGB are more advanced than RGB-D, (2) depth helps compared to pure rgb, (3) Percentage of tracked frames is LOW for all RGB-D/rgb trackers and overall peroformance is low -- room for improvement, (4) perhaps some tracking RGB mechanisms stand out and could be useful, (5) which RGB-D tracking mechanisms stand out -- depth for occlusion not enough?}

\begin{figure}[!t]
\centering
\includegraphics[width=0.95\linewidth]{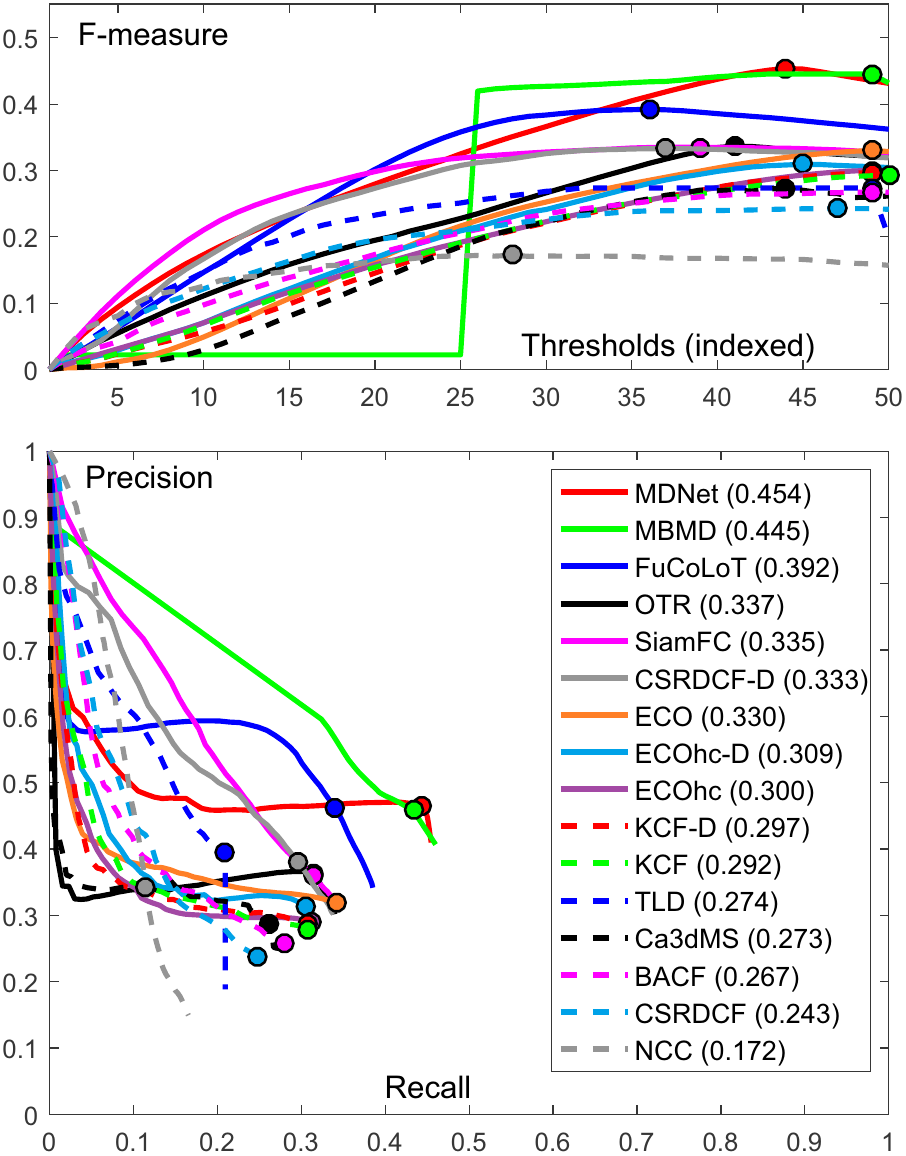}
\caption{The overall tracking performance is presented as tracking F-measure (top) and tracking Precision-Recall (bottom). Trackers are ranked by their optimal
tracking % can be deleted to optimize length
performance (maximum F-measure).}
\label{fig:overall-performance}
%\vspace{-1cm}
\end{figure}

Figure~\ref{fig:overall-performance} shows trackers ranked according to the F-measure, while tracking Precision-Recall plots are visualized for additional insights.
A striking result is that the overall top-performing trackers are pure RGB trackers, which do not use depth information at all.
MDNet and MBMD achieve comparable F-score, while FuCoLoT ranks third. It is worth mentioning that all three trackers are long-term with strong re-detection capability~\cite{Lukezic_arxiv_lt_bench}.
Even though MDNet was originally published as a short-term tracker, it has been shown that it performs well in a long-term scenario~\cite{Lukezic_arxiv_lt_bench,moudgil_lt_benchmark,valmadre_lt_benchmark} due to its powerful CNN-based classifier with selective update and hard negative mining. Another long-term tracker, TLD, is ranked very low despite its re-detection capability, due to a fairly simplistic visual model which is unable to capture complex target appearance changes.

%Another important aspect is re-detection capability since target is often not visible due to the full occlusion or out-of-frame disappearance. Therefore MBMD, which is winner of the VOT18 long-term challenge~\cite{Vot2018}, and FuCoLoT~\cite{lukezicFCLT} rank so high. Another long-term tracker, TLD, is ranked so low despite being a long-term tracker, due to its simple visual model which is not able to capture complex appearance changes of the target.

State-of-the-art RGB-D trackers, OTR and CSRDCF-D, using only hand-crafted features,
achieve a comparable performance to complex deep-features-based short-term RGB trackers ECO and SiamFC. 
This implies that modern RGB deep features may compensate for the lack of depth information to some extent. 
On the other hand, state-of-the-art RGB trackers show improvements when extended by depth channel (CSRDCF-D, ECOhc-D and KCF-D). 
This means that existing RGB-D trackers lag behind the state-of-the-art RGB trackers which is a large opportunity for improvement by utilizing deep features combined with depth information.

%The state-of-the-art RGB-D trackers, OTR and CSRDCF-D are using hand-crafted features like HoG and Colornames and they achieve comparable performance to the short-term RGB trackers ECO and SiamFC which use complex deep features. 
%However, trackers which are extended RGB trackers with depth information, e.g., ECOhc-D, CSRDCF-D and KCF-D achieve improved tracking performance. 
%These results show that existing RGB-D trackers are not powerful like state-of-the-art RGB trackers yet, but depth is a strong cue to boost tracking performance, which offers a lot of room for improvement. 
 
Overall, both state-of-the-art RGB and RGB-D trackers exhibit a relatively low performance. For example, tracking Recall can be interpreted as the average overlap with ground truth on frames in which the target is visible. This value is below 0.5 for all trackers, which implies the dataset is particularly challenging for all trackers and offers significant potential for tracker improvement. 

\begin{figure}
\centering
\includegraphics[width=\linewidth]{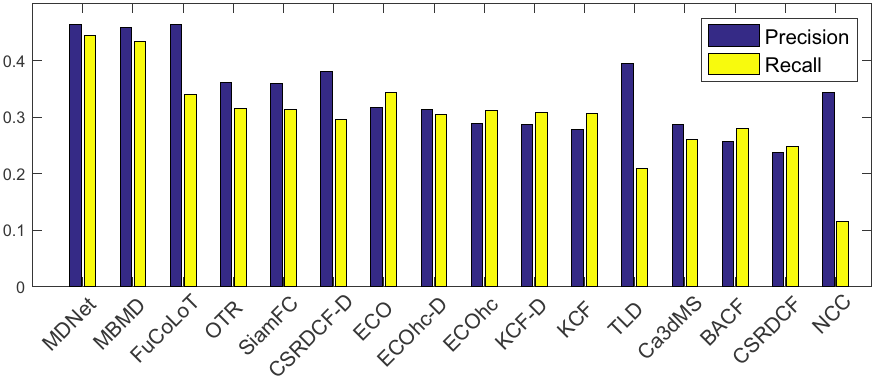}
\caption{Tracking precision and recall calculated at the optimal point (maximum F-measure).}
\label{fig:overall-precision-recall}
\end{figure} 

\begin{figure*}[h]
\centering
\includegraphics[width=\linewidth]{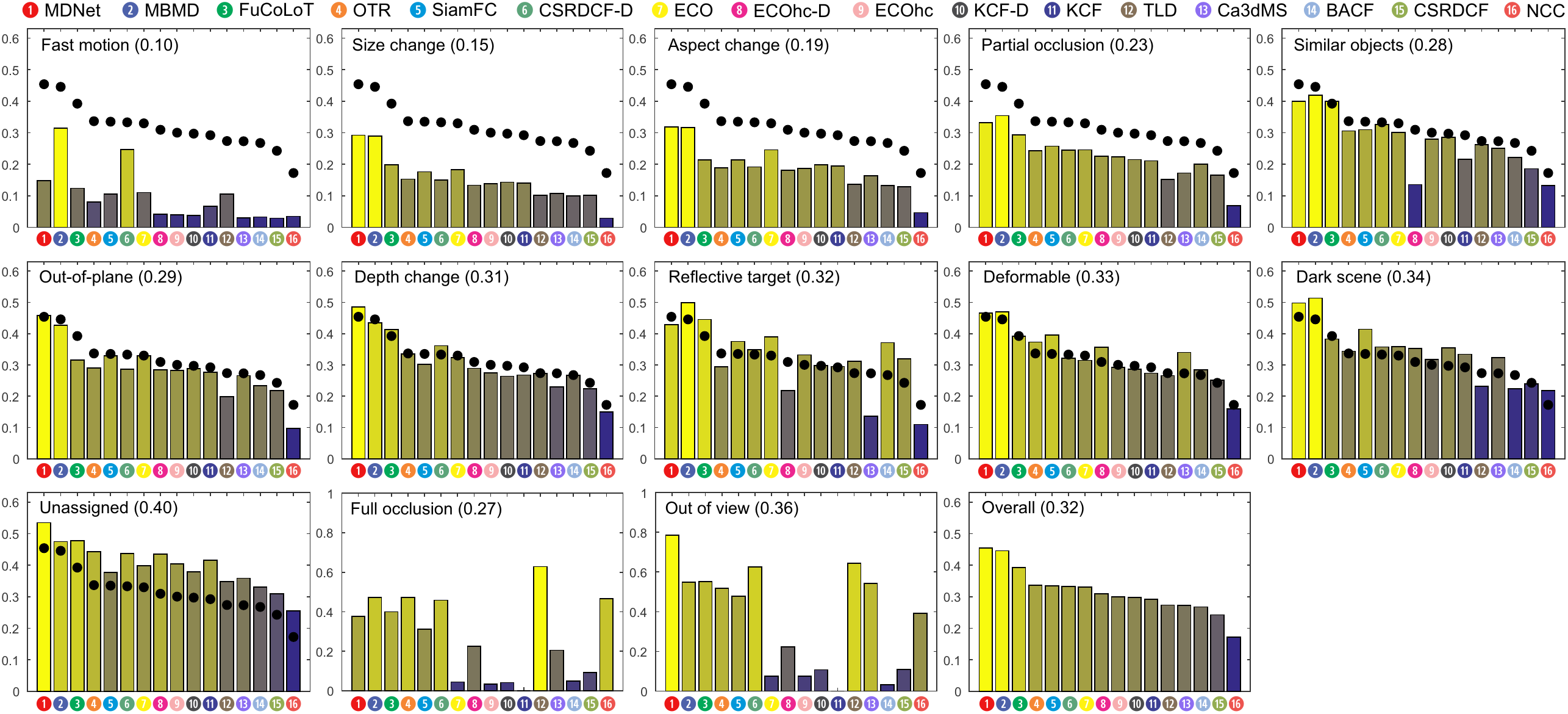}
\caption{Tracking performance w.r.t. visual attributes. The first eleven attributes correspond to scenarios with a visible target (showing F-measure). The overall tracking performance is shown in each graph with black dots.
The attributes {\it full occlusion} and {\it out of view} represent periods when the target is not visible and true negative rate is used to measure the performance.}
\label{fig:per-attribute-performance}
\end{figure*}

% \paragraph{Precision-recall analysis.}
{\bf Precision-recall analysis.}
For further performance insights, we visualize the tracking Precision and Recall at the optimal tracking point, i.e., at the highest F-measure, in Figure~\ref{fig:overall-precision-recall}. Precision and Recall  are similarly low for most trackers, implying that trackers need to improve in target detection as well as localization accuracy. FuCoLoT, CSRDCF-D and TLD obtain significantly higher Precision than Recall, which means that mechanism for reporting loss of target is rather conservative in these trackers -- a typical property we observed in all long-term trackers. The NCC tracker achieves significantly higher precision than recall, but this is a degenerated case since the target is reported as lost for most part of the sequence (very low Recall).
 
Another interesting observation is that tracking precision of the FuCoLoT is comparable to the top-performing MDNet and MBMD which shows that predictions made by FuColoT are similarly accurate to those made by top-performing trackers. 
On the other hand, top-performing MDNet and MBMD have a much higher recall, which shows that they are able to correctly track much more frames where the target is visible, which might again be attributed to the use of deep features.

% \paragraph{Overall findings.}
{\bf Overall findings.} 
We can identify several good practices in the tracking architectures that look promising according to the overall results.
Methods based on {\it deep features} show promise in capturing complex target appearance changes. We believe that {\it training} deep features on depth offers an opportunity for performance boost.
A reliable {\it failure detection mechanism} is an important property for RGB-D tracking. Depth offers a convenient cue for detection of such events and combined with image-wide {\it re-detection} some of the RGB-D trackers address the long-term tracking scenario well.
Finally, we believe that depth offers a rich {\it information complementary} to RGB for 3D target appearance modeling and depth-based target separation from the background, which can contribute in target localization. None of the existing RGB-D trackers incorporates all of these architectural elements, which opens a lot of new research opportunities.

\subsection{Per-attribute Tracking Performance}  \label{sec:per-attribute-experiment}

The trackers were also evaluated on thirteen visual attributes (Section~\ref{sec:sequence-annotation}) in Figure~\ref{fig:per-attribute-performance}. Performance on the attributes with visible target is quantified by the average F-measure, while true-negative rate (TNR~\cite{valmadre_lt_benchmark}) is used to quantify the performance under full occlusion and out-of-view target disappearance.
 
%As expected, performance on the attribute {\it unassigned} is higher than global performance on all trackers.

Performance of all trackers is very low on {\it fast-motion}, making it the most challenging attribute. The reason for performance degradation is most likely the relatively small frame-to-frame target search range. Some of the long-term RGB-D and RGB trackers, e.g., MBMD and CSRDCF-D, stand out from the other trackers due to a well-designed image-wide re-detection mechanism, which compensates for a small frame-to-frame receptive field.
 
The next most challenging attributes are target {\it size change} and {\it aspect change}. MDNet and MBMD significantly outperform the other trackers since they explicitly estimate the target aspect.
Size change is related to depth change, but the RGB-D tracker do not exploit this, which opens an opportunity for further research in depth-based robust scale adaptation.
 
{\it Partial occlusion} is particularly challenging for both RGB and RGB-D trackers. Failing to detect occlusion can lead to adaptation of the visual model to the occluding object and eventual tracking drift. In addition, too small frame-to-frame target search region leads to failure of target re-detection after the occlusion.
 
The attributes {\it similar objects}, {\it out-of-plane rotation}, {\it deformable}, {\it depth-change} and {\it dark scene} do not significantly degrade the performance compared to the overall performance. Nevertheless, the overall performance of trackers is rather low, which leaves plenty of room for improvements. We observe a particularly large drop in ECOhc-D on the similar-objects attribute which indicates that the tracker locks on to the incorrect/similar object at target re-detection stage.
 
The {\it reflective target} attribute, unique for objects such as metal cups, mostly affects RGB-D trackers. The reason is that objects of this class are fairly well distinguished from the background in RGB, while their depth image is consistently unreliable. This means that more effort should be put in information fusion part of the RGB-D trackers.
 
The attributes {\it deformable} and {\it dark-scene} are very well addressed by deep trackers (MDNet, MBMD, SiamFC and ECO), which makes them the most promising for coping with such situations.
It seems that normalization, non-linearity and pooling in CNNs make deep features sufficiently invariant to image intensity changes and object deformations observed in practice.
%\cmnt{[AL] Note that graphs for full occlusion and out-of-frame are not included yet, but they will be in Figure~\ref{fig:per-attribute-performance} next to other bar plots.}

%The attributes {\it full occlusion} and {\it out-of-frame} target disappearances correspond to the frames where target is not visible.
%Therefore an ability of the tracker to predict such events is analyzed and measured as a true negative rate.

{\it Full occlusions} are usually short-lasting events. On average, the trackers detect full a occlusion with some delay, thus a large percentage of occlusion frames are mistaken for the target visible. This implies poor ability to distinguish the appearance change due to occlusion from other appearance changes. The best target absence prediction at full occlusion is achieved by TLD, which is the most conservative in predicting target presence.

Situations when the target leaves the field of view ({\it out-of-view} attribute) are better predictable than full occlusions, due to longer target absence periods. Long-term trackers are performing very well in these situations and conservative visual model update seems to be beneficial.

\begin{figure}[ht]
\centering
\includegraphics[width=\linewidth]{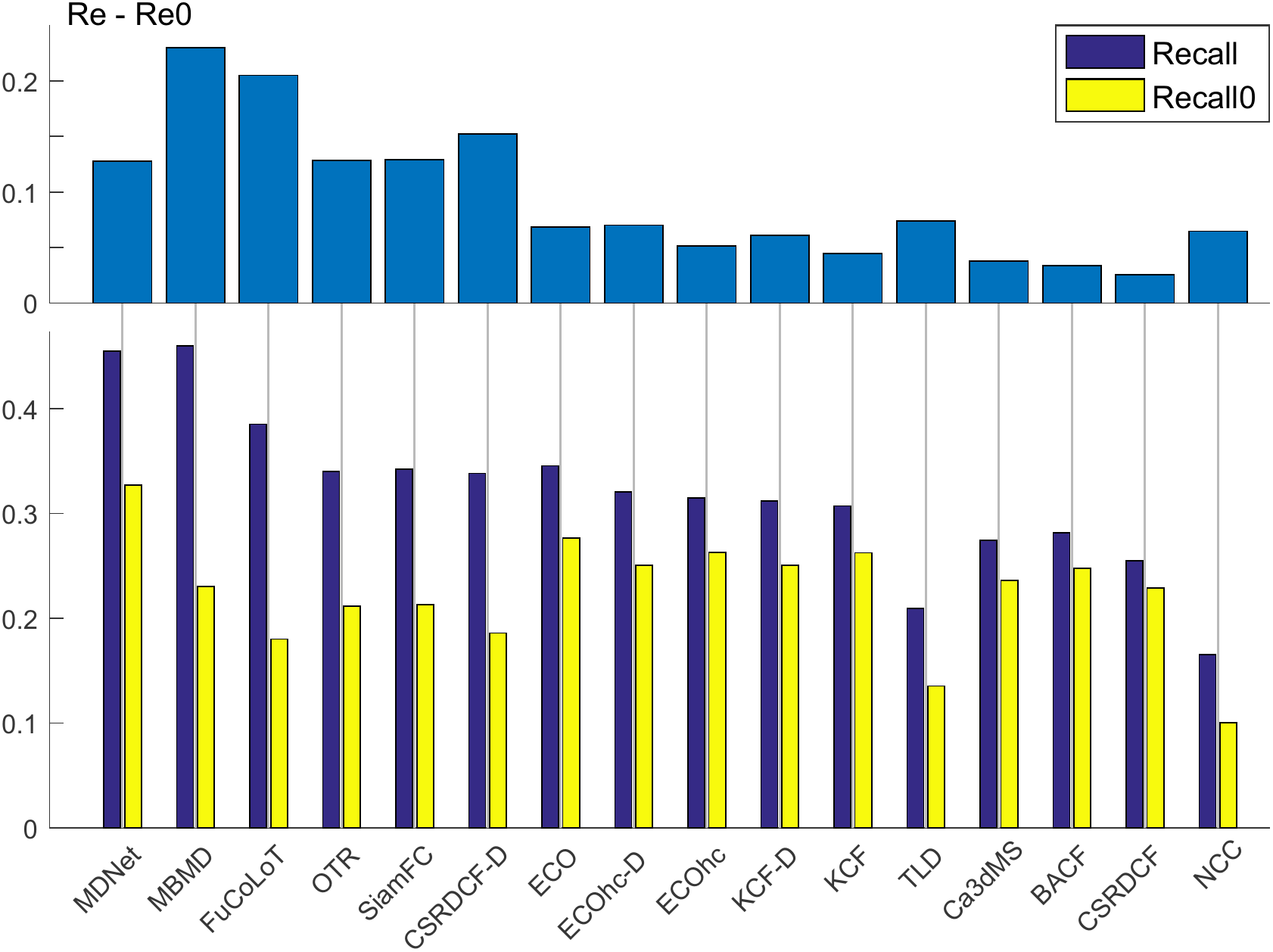}
\caption{No redetection experiment. Tracking recall is shown on the bottom graph as dark blue bars. Modified tracking recall ($Re0$) is shown as yellow bars and it is calculated by setting the per-frame overlaps to zero after the first tracking failure. 
The difference between both recalls is shown on top. A large difference indicates good re-detection capability of the tracker.}
\label{fig:no-redetection-performance}
\end{figure}

A no-redetection experiment from~\cite{Lukezic_arxiv_lt_bench} was performed to measure target re-detection capability in the considered trackers (Figure~\ref{fig:no-redetection-performance}). 
In this experiment the standard tracking Recall ($Re$) is compared to a recall ($Re_0$) computed on modified tracker output -- all overlaps are set to zero after the first occurrence of the zero overlap (i.e., the first target loss). 
Large difference between the recalls ($Re - Re_0$) indicates a good re-detection capability of a tracker. 
The trackers with the largest re-detection capability are MBMD, FuCoLoT (RGB trackers) and CSRDCF-D (RGB-D extension of CSRDCF) followed by OTR (RGB-D tracker) and two RGB trackers MDNet and SiamFc.

%The following attributes are annotated in the proposed dataset: target out-of-view, full occlusion, partial occlusion, out-of-plane rotation, fast motion, similar objects, target size change, aspect ratio change, deformable target, reflective target depth change and dark scene. 
%All attributes are annotated per-frame and frames which are not annotated with none of the attribute is marked as {\it unassigned}.
%Tracking performance in terms of tracking F-measure for each attribute is presented in Figure[REF]. 
%Comment results. Which architectures are the most promising for RGB-D tracking. Does doepth help for tracking? (here we should comment results of original vs. dpth-modified trackers). Per-property analysis: what is the most challenging attribute and why (how it could be addressed)? Where does depth NOT help (reflective objects, anything else?)

%\subsection{Influence of illumination change}  %\label{sec:illuminationchange}
%Illumination periodically reduced to $10\%$ of original illumination. Period \cmnt{xx} was used in all sequences.
 
%\subsection{Influence of Depth Quality}  \label{sec:depth-quality-influence}
%Depth + noise (random pixels to 0), Shift depth for a certain portion

%\subsection{Qualitative Analysis}  \label{sec:qualitative}
%Show images and find something interesting (support the major findings of the paper with these visual examples).

%%%%%%%%%%%%%%%%%%%%%%%%%%%%%%%%%%%%%%%%%%%%%%%%%%%%
\section{Conclusion}  \label{sec:conclusion}

We proposed a color-and-depth general visual object tracking benchmark (CDTB) that goes beyond the existing benchmarks in several ways. CDTB is the only benchmark with RGB-D dataset recorded by several color-and-depth sensors, which allows inclusion of indoor and outdoor sequences captured under unconstrained conditions (e.g., direct sun light) and covers a wide range of realistic depth signals. Empirical comparison to related datasets shows that CDTB contains a much higher level of object pose change and exceeds the other datasets in the number of frames by an order of magnitude. The objects disappear and reappear far more often, with disappearance periods ten times longer than in other benchmarks. Performance of trackers is lower on CDTB than related datasets. CDTB is thus currently the most challenging dataset, which allows RGB-D general object tracking evaluation under various realistic conditions involving target disappearance and re-appearance. 

We evaluated recent state-of-the-art (SotA) RGB-D and RGB trackers on
CDTB. Results show that SotA RGB trackers outperform SotA RGB-D trackers, which means that the architectures of RGB-D trackers could benefit from adopting (and adapting) elements of the recent RGB SotA. Nevertheless, the performance of all RGB and RGB-D trackers is rather low, leaving a significant room for improvements. 

Detailed performance analysis showed several insights. Performance of baseline RGB trackers improved already from straightforward addition of the depth information. Current mechanisms for color and depth fusion in RGB-D trackers are inefficient and perhaps deep features trained on RGB-D data should be considered. RGB-D trackers do not fully exploit the depth information for robust object scale estimation. Fast motion is particularly challenging for all trackers indicating that short-term target search ranges should be increased. Target detection and mechanisms for detecting target loss have to be improved as well. We believe these insights in combination with the presented benchmark will spark further advancements in RGB-D tracking and contribute to closing the gap between RGB and RGB-D state-of-the-art.
 
%%%%%%%%%%%%%%%%%%%%%%%%%%%%%%%%%%%%%%%%%%%%%%%%%%%%
%%%%%%%%%%%%%%%%%%%%%%%%%%%%%%%%%%%%%%%%%%%%%%%%%%%%
{\small
\bibliographystyle{ieee}
\bibliography{bib}

\begin{thebibliography}{10}\itemsep=-1pt

\bibitem{DLST}
N.~An, X.-G. Zhao, and Z.-G. Hou.
\newblock {Online RGB-D Tracking via Detection-Learning-Segmentation}.
\newblock In {\em ICPR}, 2016.

\bibitem{bertinetto2016fully}
L.~Bertinetto, J.~Valmadre, J.~F. Henriques, A.~Vedaldi, and P.~H.~S. Torr.
\newblock {Fully-Convolutional Siamese Networks for Object Tracking}.
\newblock In {\em ECCV Workshops}, 2016.

\bibitem{Bibi3D}
A.~Bibi, T.~Zhang, and B.~Ghanem.
\newblock {3D Part-Based Sparse Tracker with Automatic Synchronization and
  Registration}.
\newblock In {\em CVPR}, 2016.

\bibitem{Bolme-2010-cvpr}
D.~S. Bolme, J.~Beveridge, B.~A. Draper, and Y.-M. Lui.
\newblock {Visual Object Tracking using Adaptive Correlation Filters}.
\newblock In {\em CVPR}, 2010.

\bibitem{Buch-2013-icra}
A.~Buch, D.~Kraft, J.-K. Kamarainen, H.~Petersen, and N.~Kruger.
\newblock {Pose estimation using local structure-specific shape and appearance
  context}.
\newblock In {\em ICRA}, 2013.

\bibitem{dskcf_bmvc}
M.~Camplani, S.~Hannuna, M.~Mirmehdi, D.~Damen, A.~Paiement, L.~Tao, and
  T.~Burghardt.
\newblock {Real-time RGB-D Tracking with Depth Scaling Kernelised Correlation
  Filters and Occlusion Handling}.
\newblock In {\em BMVC}, 2015.

\bibitem{choi13iros_rgbdtracking}
C.~Choi and H.~Christensen.
\newblock {RGB}-d object tracking: {A} particle filter approach on {GPU}.
\newblock In {\em IROS}, 2013.

\bibitem{choi_pami13}
W.~Choi, C.~Pantofaru, and S.~Savarese.
\newblock {A General Framework for Tracking Multiple People from a Moving
  Camera}.
\newblock {\em IEEE PAMI}, 2013.

\bibitem{HOG}
N.~Dalal and B.~Triggs.
\newblock {Histograms of Oriented Gradients for Human Detection}.
\newblock In {\em CVPR}, 2005.

\bibitem{ECO}
M.~Danelljan, G.~Bhat, F.~Shahbaz~Khan, and M.~Felsberg.
\newblock {ECO: Efficient Convolution Operators for Tracking}.
\newblock In {\em CVPR}, 2017.

\bibitem{Ess-2008-cvpr}
A.~Ess, B.~Leibe, K.~Schindler, , and L.~van Gool.
\newblock {A Mobile Vision System for Robust Multi-Person Tracking}.
\newblock In {\em CVPR}, 2008.

\bibitem{Galoogahi-2015-cvpr}
H.~Galoogahi, T.~Sim, and S.~Lucey.
\newblock {Correlation Filters with Limited Boundaries}.
\newblock In {\em CVPR}, 2015.

\bibitem{Garcia-Hernando-2018-cvpr}
G.~Garcia-Hernando, S.~Yuan, S.~Baek, and T.-K. Kim.
\newblock {First-Person Hand Action Benchmark with RGB-D Videos and 3D Hand
  Pose Annotations}.
\newblock In {\em CVPR}, 2018.

\bibitem{Hannuna2016}
S.~Hannuna, M.~Camplani, J.~Hall, M.~Mirmehdi, D.~Damen, T.~Burghardt,
  A.~Paiement, and L.~Tao.
\newblock {DS-KCF: A Real-time Tracker for RGB-D Data}.
\newblock {\em Journal of Real-Time Image Processing}, 2016.

\bibitem{HZ}
R.~I. Hartley and A.~Zisserman.
\newblock {\em Multiple View Geometry in Computer Vision}.
\newblock Second edition, 2004.

\bibitem{Henriques_KCF}
J.~F. Henriques, R.~Caseiro, P.~Martins, and J.~Batista.
\newblock {High-Speed Tracking with Kernelized Correlation Filters}.
\newblock {\em IEEE PAMI}, 37(3):583--596, 2015.

\bibitem{Hirschmuller2005}
H.~Hirschmuller.
\newblock {Accurate and Efficient Stereo Processing by Semi-Global Matching and
  Mutual Information}.
\newblock In {\em CVPR}, 2005.

\bibitem{TLD}
Z.~Kalal, K.~Mikolajczyk, and J.~Matas.
\newblock {Tracking-Learning-Detection}.
\newblock {\em IEEE PAMI}, 34(7):1409--1422, 2011.

\bibitem{Kart_ECCVW}
U.~{Kart}, J.-K. {K{\"a}m{\"a}r{\"a}inen}, and J.~{Matas}.
\newblock {How to Make an RGBD Tracker ?}
\newblock In {\em ECCV Workshops}, 2018.

\bibitem{DMDCF}
U.~{Kart}, J.-K. {K{\"a}m{\"a}r{\"a}inen}, J.~{Matas}, L.~{Fan}, and
  F.~{Cricri}.
\newblock {Depth Masked Discriminative Correlation Filter}.
\newblock In {\em ICPR}, 2018.

\bibitem{OTR}
U.~{Kart}, A.~Luke{\v{z}}i{\v{c}}, M.~Kristan, J.-K. {K{\"a}m{\"a}r{\"a}inen},
  and J.~{Matas}.
\newblock {Object Tracking by Reconstruction with View-Specific Discriminative
  Correlation Filters}.
\newblock In {\em CVPR}, 2019.

\bibitem{Galoogahi_2017_ICCV}
H.~Kiani~Galoogahi, A.~Fagg, and S.~Lucey.
\newblock {Learning Background-Aware Correlation Filters for Visual Tracking}.
\newblock In {\em ICCV}, 2017.

\bibitem{vot2016}
M.~Kristan, A.~Leonardis, J.~Matas, M.~Felsberg, R.~Pflugfelder,
  L.~{\v{C}}ehovin, T.~Voj{\'{\i}}r, and et~al.
\newblock {The Visual Object Tracking VOT2016 Challenge Results}.
\newblock In {\em ECCV Workshops}, 2016.

\bibitem{vot2017}
M.~Kristan, A.~Leonardis, J.~Matas, M.~Felsberg, R.~Pflugfelder, and et~al.
\newblock {The Visual Object Tracking VOT2017 Challenge Results}.
\newblock In {\em ICCV Workshops}, 2017.

\bibitem{Vot2018}
M.~Kristan, A.~Leonardis, J.~Matas, M.~Felsberg, R.~Pfugfelder, L.~C. Zajc, and
  T.~V. et~al.
\newblock {The sixth Visual Object Tracking VOT2018 challenge results}.
\newblock In {\em ECCV Workshops}, 2018.

\bibitem{vot2015}
M.~Kristan, J.~Matas, A.~Leonardis, M.~Felsberg, and L.~e.~a.
  {\v{C}}ehovin~Zajc.
\newblock {The Visual Object Tracking VOT2015 Challenge Results}.
\newblock In {\em ICCV Workshops}, 2015.

\bibitem{Kristan2016Pami}
M.~Kristan, J.~Matas, G.~Nebehay, F.~Porikli, and L.~{\v{C}}ehovin.
\newblock {A Novel Performance Evaluation Methodology for Single-Target
  Trackers}.
\newblock {\em IEEE PAMI}, 38(11):2137--2155, 2016.

\bibitem{Vot2014}
M.~Kristan, R.~Pflugfelder, A.~Leonardis, J.~Matas, L.~{\v{C}}ehovin,
  G.~Nebehay, T.~Voj{\'{\i}}r, and et~al.
\newblock {The Visual Object Tracking VOT2014 Challenge Results}.
\newblock In {\em ECCV Workshops}, 2014.

\bibitem{Vot2013}
M.~Kristan, R.~Pflugfelder, A.~Leonardis, J.~Matas, F.~Porikli, and et~al.
\newblock {The Visual Object Tracking VOT2013 Challenge Results}.
\newblock In {\em CVPR Workshops}, 2013.

\bibitem{ca3dms}
Y.~Liu, X.-Y. Jing, J.~Nie, H.~Gao, J.~Liu, and G.-P. Jiang.
\newblock {Context-aware 3-D Mean-shift with Occlusion Handling for Robust
  Object Tracking in RGB-D Videos}.
\newblock {\em IEEE TMM}, 2018.

\bibitem{Fucolot}
A.~Luke{\v{z}}i{\v{c}}, L.~{\v{C}}ehovin~Zajc, T.~Voji{\v{r}}, J.~Matas, and
  M.~Kristan.
\newblock {FuCoLoT - A Fully-Correlational Long-Term Tracker}.
\newblock In {\em ACCV}, 2018.

\bibitem{csr}
A.~Luke{\v{z}}i{\v{c}}, T.~Voj{\'{\i}}r, L.~{\v{C}}ehovin, J.~Matas, and
  M.~Kristan.
\newblock {Discriminative Correlation Filter with Channel and Spatial
  Reliability}.
\newblock In {\em CVPR}, 2017.

\bibitem{Lukezic_arxiv_lt_bench}
A.~Lukezic, L.~C. Zajc, T.~Voj{\'{\i}}r, J.~Matas, and M.~Kristan.
\newblock {Now you see me: evaluating performance in long-term visual
  tracking}.
\newblock {\em CoRR}, abs/1804.07056, 2018.

\bibitem{MESHGI_OAPF}
K.~Meshgi, S.~ichi Maeda, S.~Oba, H.~Skibbe, Y.~zhe Li, and S.~Ishii.
\newblock {An Occlusion-aware Particle Filter Tracker to Handle Complex and
  Persistent Occlusions}.
\newblock {\em CVIU}, 150:81 -- 94, 2016.

\bibitem{moudgil_lt_benchmark}
A.~Moudgil and V.~Gandhi.
\newblock {Long-Term Visual Object Tracking Benchmark}.
\newblock In {\em ACCV}, 2018.

\bibitem{uav_benchmark_simulator}
M.~Mueller, N.~Smith, and B.~Ghanem.
\newblock {A Benchmark and Simulator for UAV Tracking}.
\newblock In {\em ECCV}, 2016.

\bibitem{Muller_2018_ECCV}
M.~Muller, A.~Bibi, S.~Giancola, S.~Alsubaihi, and B.~Ghanem.
\newblock {TrackingNet: A Large-Scale Dataset and Benchmark for Object Tracking
  in the Wild}.
\newblock In {\em ECCV}, 2018.

\bibitem{MDNet}
H.~Nam and B.~Han.
\newblock {Learning Multi-Domain Convolutional Neural Networks for Visual
  Tracking}.
\newblock In {\em CVPR}, 2016.

\bibitem{Richter-2016-eccv}
S.~Richter, V.~Vineet, S.~Roth, and V.~Koltun.
\newblock {Playing for Data: Ground Truth from Computer Games}.
\newblock In {\em ECCV}, 2016.

\bibitem{smeulders_pami_2014}
A.~W.~M. Smeulders, D.~M. Chu, R.~Cucchiara, S.~Calderara, A.~Dehghan, and
  M.~Shah.
\newblock {Visual Tracking: An Experimental Survey}.
\newblock {\em IEEE PAMI}, 36(7):1442--1468, 2014.

\bibitem{princetonrgbd}
S.~Song and J.~Xiao.
\newblock {Tracking Revisited Using RGBD Camera: Unified Benchmark and
  Baselines}.
\newblock In {\em ICCV}, 2013.

\bibitem{Spinello-2011-iros}
L.~Spinello and K.~O. Arras.
\newblock {People detection in {RGB-D} data}.
\newblock In {\em IROS}, 2011.

\bibitem{valmadre_lt_benchmark}
J.~Valmadre, L.~Bertinetto, J.~F. Henriques, R.~Tao, A.~Vedaldi, A.~W.~M.
  Smeulders, P.~H.~S. Torr, and E.~Gavves.
\newblock {Long-term Tracking in the Wild: A Benchmark}.
\newblock In {\em ECCV}, 2018.

\bibitem{OTB}
Y.~Wu, J.~Lim, and Y.~Ming-Hsuan.
\newblock {Object Tracking Benchmark}.
\newblock {\em IEEE PAMI}, 37:1834 -- 1848, 2015.

\bibitem{STC}
J.~Xiao, R.~Stolkin, Y.~Gao, and A.~Leonardis.
\newblock {Robust Fusion of Color and Depth Data for RGB-D Target Tracking
  Using Adaptive Range-Invariant Depth Models and Spatio-Temporal Consistency
  Constraints}.
\newblock {\em IEEE Transactions on Cybernetics}, 48:2485 -- 2499, 2018.

\bibitem{MBMD}
Y.~Zhang, D.~Wang, L.~Wang, J.~Qi, and H.~Lu.
\newblock {Learning Regression and Verification Networks for Long-term Visual
  Tracking}.
\newblock {\em CoRR}, abs/1809.04320, 2018.

\end{thebibliography}
}

\end{document}